# Optimizing Filter Size in Convolutional Neural Networks for Facial Action Unit Recognition

Shizhong Han[1], Zibo Meng[1], Zhiyuan Li[1], James O'Reilly[1], Jie Cai[1], Xiaofeng Wang[2], Yan Tong[1]
[1]Department of Computer Science & Engineering
[2]Department of Electrical Engineering
University of South Carolina, Columbia, SC
`han38, mengz, oreillyj, jcai, wangxi, tongy@email.sc.edu`

## Abstract

*Recognizing facial action units (AUs) during spontaneous facial displays is a challenging problem. Most recently, Convolutional Neural Networks (CNNs) have shown promise for facial AU recognition, where predefined and fixed convolution filter sizes are employed. In order to achieve the best performance, the optimal filter size is often empirically found by conducting extensive experimental validation. Such a training process suffers from expensive training cost, especially as the network becomes deeper.*

*This paper proposes a novel Optimized Filter Size CNN (OFS-CNN), where the filter sizes and weights of all convolutional layers are learned simultaneously from the training data along with learning convolution filters. Specifically, the filter size is defined as a continuous variable, which is optimized by minimizing the training loss. Experimental results on two AU-coded spontaneous databases have shown that the proposed OFS-CNN is capable of estimating optimal filter size for varying image resolution and outperforms traditional CNNs with the best filter size obtained by exhaustive search. The OFS-CNN also beats the CNN using multiple filter sizes and more importantly, is much more efficient during testing with the proposed forward-backward propagation algorithm.*

## 1. Introduction

Facial behavior is a natural and powerful means for human communications. Facial Action Coding System (FACS) developed by Ekman and Friesen [6] describes facial behavior with a set of facial action units (AUs), each of which is anatomically related to the contraction of a set of facial muscles. An automatic AU recognition system has various applications in human-computer interaction (HCI) such as interactive games, advertisement impact analysis, and synthesizing human expression. However, it is still a challenging problem to recognize facial AUs from spontaneous facial displays, especially with large variations in facial appearance caused by free head movements, occlusions, and illumination changes.

Extensive efforts have been focused on extracting features that are capable of capturing facial appearance and/or geometrical changes caused by AUs. While most of the earlier approaches employed handcrafted and general-purposed features; deep learning, especially CNN based methods, has shown great promise in recognizing facial expressions or AUs [7, 24, 19, 15, 9, 12, 34, 17, 30, 21].

In CNNs, the size of the convolution filters determines the size of receptive field where information is extracted. CNN-based methods employ predefined and fixed filter sizes in each convolutional layer, which is called the *traditional CNN* hereafter. In general, larger filter sizes are employed in the lower convolutional layers, whereas smaller filter sizes are used in the upper layers [18, 4]. However, the fixed filter sizes are not necessarily optimal for all applications/tasks as well as for different image resolutions. Specifically, different AUs cause facial appearance changes over various regions at different scales and therefore, may prefer different filter sizes. For example, *long* and deep nasolabial furrows are important for recognizing AU10 (upper lip raiser), while *short* "wrinkles in the skin above and below the lips" and small bulges below the lower lip are cues for recognizing AU23 (lip tightener) [6].

Given a predefined input image size, the best filter size is often selected experimentally or by visualization [32] for each convolutional layer. For example, Kim et al. [17], who achieved the best expression recognition performance of EmotiW2015 challenge [5], experimentally selected the best filter sizes for the three convolutional layers. *However, with CNNs becoming deeper and deeper [23, 11], it is impractical to search for the best filter size by exhaustive search, due to the highly expensive training cost.*

In this work, we propose a novel and feasible solution in a CNN framework to automatically learn the filter sizes for all convolutional layers simultaneously from the training data along with learning the convolution filters. In particular, we proposed an Optimized Filter Size CNN (OFS-CNN), where the optimal filter size of each convolutional layer is estimated iteratively using stochastic gradient descent (SGD) during the *backpropagation process*. As illustrated in Figure. 1, the filter size $k$ of a convolutional layer, which is a constant in the traditional CNNs, is defined as a continuous variable in the OFS-CNN. During backpropagation, the filter size $k$ will be updated, e.g., decreased when the partial derivative of CNN loss with respect to the filter size is positive, i.e., $\frac{\partial L}{\partial k} > 0$, and vice versa.

In this work, a forward-backward propagation algorithm is proposed to estimate the filter size iteratively. To facilitate the convolution operation with a continuous filter size, *upper-bound* and *lower-bound* filters with integer-sizes are defined. In the *forward process*, an activation resulted from a convolution operation with a continuous filter size can be calculated as the interpolation of the activations using the upper-bound and lower-bound filters. Furthermore, we show that only one convolution operation is needed with the upper-bound and lower-bound filters. Therefore, *the proposed OFS-CNN has similar computation complexity as the traditional CNNs in the forward process as well as in the testing process*. During *backpropagation*, the partial derivative of the activation with respect to the filter size $k$ is defined, from which $\frac{\partial L}{\partial k}$ can be calculated. With a change in the filter size $k$, the filter sizes of the upper-bound or lower-bound filters may be updated via a *transformation operation* proposed in this work.

Experimental results on two benchmark AU-coded spontaneous databases, i.e., FERA2015 BP4D database [26] and Denver Intensity of Spontaneous Facial Action (DISFA) database [20] have demonstrated that the proposed OFS-CNN outperforms the traditional CNNs with the best filter size obtained by exhaustive search and achieves state-of-the-art performance for AU recognition. Furthermore, the OFS-CNN also beats a deep CNN using multiple filter sizes with a remarkable improvement in time efficiency during testing, which is highly desirable for realtime applications. In addition, the OFS-CNN is capable of estimating optimal filter size for varying image resolution.

## 2. Related Work

Extensive efforts have been devoted to extracting the most effective features that characterize facial appearance and geometry changes caused by activation of facial expressions or AUs. The earlier approaches adopted various handcrafted features such as Gabor wavelets [3], histograms of Local Binary Patterns (LBP) [27], Histogram of Oriented Gradients (HOG) [2], Scale Invariant Feature Transform (SIFT) features [31], histograms of Local Phase Quantization (LPQ) [14], and their spatiotemporal extensions [14, 33, 29].

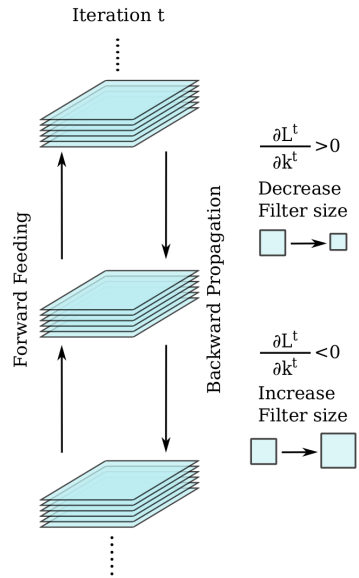

Figure 1. An overview of the proposed method to optimize the convolution filter size $k$ with the CNN loss backpropagation at the $t^{th}$ iteration. $\frac{\partial L^t}{\partial k^t}$ is the partial derivative of the loss with respect to the filter size at the $t^{th}$ iteration ($k^t$). The filter size $k$ will decrease when $\frac{\partial L^t}{\partial k^t} > 0$, and vice versa.

Most recently, CNNs have attracted increasing attention and shown great promise for facial expression and AU recognition [7, 24, 19, 15, 9, 12, 34, 17, 30, 21, 28, 25]. For example, the top 3 methods [17, 30, 21] in the recent EmotiW2015 challenge [5] are all based on CNNs and have been demonstrated to be more robust to real world conditions for facial expression recognition. All those CNN-based methods use fixed-size convolution filters.

To achieve the best performance, the optimal filter size is usually chosen empirically by either experimental validation or visualization for each convolutional layer [32]. For example, Kim et al. [17] experimentally compared facial expression recognition performance using different filter sizes and found that the CNN with 5×5, 4×4, and 5×5 filter sizes in the three convolutional layer, respectively, has the best performance on 42×42 input images. Zeiler and Fergus [32] found that 7×7 filters can capture more distinctive features than 11×11 filters on ImageNet dataset through visualization. However, such empirically selected filter sizes may not be optimal for all applications as well as for different image resolutions. Furthermore, it is impossible to perform an exhaustive search for the optimal combination of filter sizes of all convolutional layers for deep CNNs.

To achieve scale invariance, CNNs with multiple filter sizes have been developed. The inception module [23] concatenates the activation feature maps from 1×1, 3×3, and 5×5 filters. The Neural Fabrics [22] embeds an exponentially large number of architectures with 3 × 3 filters. Multi-grid Neural Architecture [16] concatenates the feature maps activated by pyramid filters. However, all those methods are still based on fixed filter size and more importantly, demand a significant increase in the time and space complexity due to the complex model structure.

In contrast, the proposed OFS-CNN is capable of learning and optimizing the filter sizes for all convolutional layers simultaneously in a CNN learning framework, which is desirable, especially when the CNNs go deeper and deeper. Furthermore, we show that only one convolution operation is needed in the proposed forward-backward propagation algorithm. Thus, the proposed OFS-CNN has similar computational complexity as the traditional CNNs and thus, is more efficient than the structures using multiple filter sizes.

## 3. Methodology

In this work, we propose an OFS-CNN, which is capable of optimizing and learning the filter size $k$ from the training data. In the following, we will first give a brief review of the CNN, especially the convolutional layer, and then present the forward and backward propagation processes of the OFS-CNN.

### 3.1. A Brief Review of CNNs

A CNN consists of a stack of layers such as convolutional layers, pooling layers, rectification layers, fully connected (FC) layers, and loss layers. These layers transform the input data to highly nonlinear representations. Convolutional layers are used to perform convolution on input images or feature maps from the previous layer with filters. Generally, the first convolutional layer is used to extract low-level image features such as edges; while the upper layers can extract complex and task-related features.

Given an input image/feature map denoted by $\mathbf{x}$, an activation at the $i^{th}$ row and the $j^{th}$ column, denoted by $y_{ij}$, in a convolutional layer can be calculated using the convolution operation by computing the inner product of the filter and the input as follows:

$$y_{ij}(k) = \mathbf{w}(k)^\top \mathbf{x}_{ij}(k) + b_{ij} \quad (1)$$

where $\mathbf{w}(k)$ is a convolution filter with the filter size $k \times k$; $\mathbf{x}_{ij}(k)$ denotes the input with a $k \times k$ receptive field centered at the $i^{th}$ row and the $j^{th}$ column; and $b_{ij}$ is a bias. Traditionally, the filter size $k$ is a predefined integer and fixed throughout the training/testing process. *In this work, $k \in \mathbb{R}^+$ is defined as a continuous variable that can be learned and optimized during CNN training.*

### 3.2. Forward Processing of the OFS-CNN

In the forward process, convolution operations are conducted to calculate activations using learned filters as in Eq. 1. However, the convolution operation can only be performed with integral size filters in the CNN.

**Upper-bound and lower-bound filters:** In order to build the relationship between the activation $y_{ij}$ and the continuous filter size $k$, we first define an *upper-bound filter* denoted by $\mathbf{w}(k_+)$ and a *lower-bound filter* denoted by $\mathbf{w}(k_-)$. Specifically, $k_+$ is the upper-bound filter size and is the smallest odd number that is bigger than $k$; while $k_-$ is the lower-bound filter size and is the largest odd number that is less than or equal to $k$. $k_+$ and $k_-$ can be calculated as

$$k_+ = \lfloor \frac{k+1}{2} \rfloor * 2 + 1, \quad k_- = \lfloor \frac{k+1}{2} \rfloor * 2 - 1 \quad (2)$$

Then, the activation $y_{ij}(k)$ can be defined as the linear interpolation of the activations of the upper-bound and lower-bound filters denoted by $y_{ij}(k_-)$ and $y_{ij}(k_+)$, respectively:

$$y_{ij}(k) = \alpha y_{ij}(k_+) + (1-\alpha) y_{ij}(k_-) \quad (3)$$

where $y_{ij}(k_+)$ and $y_{ij}(k_-)$ are calculated as in Eq. 1 with the same bias, but with the upper-bound and lower-bound filters, i.e., $\mathbf{w}(k_+)$ and $\mathbf{w}(k_-)$, respectively. $\alpha = \frac{(k-k_-)}{2}$ is the linear interpolation weight.

**Remark 1.** *A cubic interpolation can also be used to build the relationship between the activation $y_{ij}$ and the continuous variable $k$. However, it requires a higher computational complexity and needs at least three points; while the linear interpolation only needs two points $k_-$ and $k_+$.*

**Remark 2.** *The filter size $k$ is actually a weight-related filter size in the interval $[k_-, k_+]$ and can be calculated as:*
$$k = k_- + 2\alpha \quad (4)$$

**Convolution with a continuous filter size:** As in Remark 2, we can explicitly define the filter $\mathbf{w}(k)$ with a continuous size $k$. As shown in Fig. 2, the upper-bound and lower-bound filters are defined to share the same coefficients in the region with green color and to differ by the pink boundary denoted by $\triangle \mathbf{w}(k_+)$. Let $\triangle \mathbf{w}(k_+) = \mathbf{w}(k_+) - \mathbf{w}(k_-)$ be the ring boundary with zeros inside as shown in Fig. 2, then the filter $\mathbf{w}(k)$ with a continuous size $k$ can be defined as follows:

$$\mathbf{w}(k) = \alpha \triangle \mathbf{w}(k_+) + \mathbf{w}(k_-), \quad (5)$$

**Remark 3.** *In Eq. 5, $\mathbf{w}(k)$ and $\mathbf{w}(k_-)$ have an actual filter size of $k_+$; while $\mathbf{w}(k_-)$ is zero-padded.*

**Lemma 1.** *Given the definition of the filter $\mathbf{w}(k)$ as in Eq. 5, the activation $y_{ij}(k)$ in Eq. 3 can be simplified as:*
$$y_{ij}(k) = \mathbf{w}(k)^\top \mathbf{x}_{ij}(k_+) + b_{ij} \quad (6)$$

*Proof.* Eq. 6 can be deduced from Eq. 3 as follows:
$$\begin{aligned} y_{ij}(k) &= \alpha y_{ij}(k_+) + (1-\alpha) y_{ij}(k_-) \\ &= \alpha \mathbf{w}(k_+)^\top \mathbf{x}_{ij}(k_+) + (1-\alpha) \mathbf{w}(k_-)^\top \mathbf{x}_{ij}(k_-) + b_{ij} \end{aligned} \quad (7)$$

After padding zeros for $\mathbf{w}(k_-)$, $\mathbf{w}(k_-)^\top \mathbf{x}(k_-)$ is equivalent to $\mathbf{w}(k_-)^\top \mathbf{x}(k_+)$. Then, Eq. 7 can be simplified as follows:

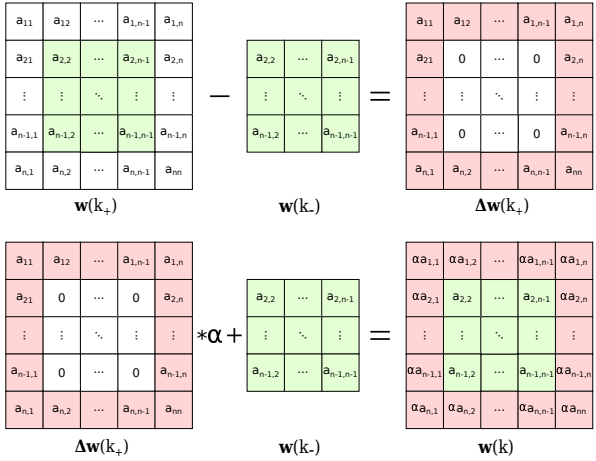

Figure 2. An illustrative definition of a filter with a continuous filter size $k \in \mathbb{R}^+$. $\mathbf{w}(k_+)$ and $\mathbf{w}(k_-)$ are the upper-bound and lower-bound filters, respectively, and share the same elements in the green region. The pink region $\triangle \mathbf{w}(k_+)$ denotes the difference between the upper-bound and lower-bound filters and has a ring shape with zeros inside. $\alpha$ is the linear interpolation weight associated with the upper-bound filter $\mathbf{w}(k_+)$. $\mathbf{w}(k)$ is a weight-related filter with a continuous filter size $k$.

$$y_{ij}(k) = \alpha \mathbf{w}(k_+)^\top \mathbf{x}(k_+) + (1-\alpha)\mathbf{w}(k_-)^\top \mathbf{x}(k_+) + b_{ij}$$
$$= \left[\alpha \mathbf{w}(k_+)^\top + (1-\alpha)\mathbf{w}(k_-)^\top\right]\mathbf{x}(k_+) + b_{ij}$$
$$= \left[\alpha \triangle \mathbf{w}(k^+)^\top + \mathbf{w}(k_-)^\top\right]\mathbf{x}(k_+) + b_{ij} \quad (8)$$

By substituting Eq. 5 into Eq. 8, we have
$$y_{ij}(k) = \mathbf{w}(k)^\top \mathbf{x}_{ij}(k_+) + b_{ij} \quad (9)$$

Thus, the activation of $y_{ij}(k)$ can be simplified as Eq. 6. □

**Remark 4.** *According to Eq. 6, only one convolution operation needs to be performed to calculate each activation $y_{ij}(k)$. Therefore, the time complexity does not increase compared with the traditional CNN in the forward training process as well as in the testing process.*

### 3.3. Backward propagation of the OFS-CNN

#### 3.3.1 Optimizing filter size in the OFS-CNN

**Calculating the partial derivative:** Since the relationship between the activation and the filter size has been defined as in Eq. 3, the partial derivative of the activation $y_{ij}$ w.r.t. the filter size can be calculated based on the derivative definition as follows:
$$\frac{\partial y_{ij}(k)}{\partial k} = \lim_{\triangle k \to 0} \frac{y_{ij}(k+\triangle k) - y_{ij}(k-\triangle k)}{2\triangle k} \quad (10)$$

When $k + \triangle k$ and $k - \triangle k$ are in the interval $[k_-, k_+]$, the derivative of each point $\frac{\partial y_{ij}(k)}{\partial k}$ is equal to the gradient of the line because of the linear interpolation. Hence, the partial derivative can be calculated as follows:

$$\frac{\partial y_{ij}(k)}{\partial k} = \frac{y_{ij}(k_+) - y_{ij}(k_-)}{k_+ - k_-} \quad (11)$$

Substituting Eq. 1 into Eq. 11, we have
$$\frac{\partial y_{ij}(k)}{\partial k} = \frac{\mathbf{w}(k_+)^\top \mathbf{x}_{ij}(k_+) - \mathbf{w}(k_-)^\top \mathbf{x}_{ij}(k_-)}{k_+ - k_-} \quad (12)$$

By padding zeros for $\mathbf{w}(k_-)$, we can simplify Eq. 12 as
$$\frac{\partial y_{ij}(k)}{\partial k} = \frac{\mathbf{w}(k_+)^\top \mathbf{x}_{ij}(k_+) - \mathbf{w}(k_-)^\top \mathbf{x}_{ij}(k_+)}{k_+ - k_-}$$
$$= \frac{\left[\mathbf{w}(k_+)^\top - \mathbf{w}(k_-)^\top\right]\mathbf{x}_{ij}(k_+)}{k_+ - k_-}$$
$$= \frac{\triangle \mathbf{w}(k_+)^\top \mathbf{x}_{ij}(k_+)}{k_+ - k_-} \quad (13)$$

Based on Eq. 13, the partial derivative of the loss $L$ w.r.t. $k$ can be calculated as follows with chain rule:
$$\frac{\partial L}{\partial k} = \sum_{i,j} \frac{\partial L}{\partial y_{ij}} \frac{\partial y_{ij}}{\partial k} \quad (14)$$

**Updating the filter size:** Given the partial derivative of the loss $L$ w.r.t. $k$, the filter size $k$ can be updated iteratively with the SGD strategy for the $(t+1)^{th}$ iteration as follows:
$$k^{t+1} = k^t - \gamma \frac{\partial L^t}{\partial k^t} \quad (15)$$

where $\gamma$ is the learning rate.

#### 3.3.2 Updating convolution filters w(k)

**Updating the upper-bound and lower-bound filters:** Since the lower-bound filter $\mathbf{w}^t(k_-)$ is defined as the inner part of the upper-bound filter $\mathbf{w}^t(k_+)$, we only need to perform backpropagation for the upper-bound filter $\mathbf{w}^t(k_+)$, which can be divided into two parts as $\mathbf{w}^t(k_+) = \mathbf{w}^t(k_-) + \triangle \mathbf{w}^t(k_+)$, where $\triangle \mathbf{w}^t(k_+)$ is the ring boundary with zeros inside and $\mathbf{w}(k_-)$ is padded with zeros. Then, the forward activation function in Eq. 6 can be reorganized as:
$$y_{ij}^t(k^t) = \mathbf{w}^t(k^t)^\top \mathbf{x}_{ij}^t(k_+^t) + b_{ij}^t$$
$$= \left[\alpha^t \triangle \mathbf{w}^t(k_+^t)^\top + \mathbf{w}^t(k_-^t)^\top\right]\mathbf{x}_{ij}^t(k_+^t) + b_{ij}^t$$
$$= \alpha^t \triangle \mathbf{w}(k_+^t)^\top \triangle \mathbf{x}_{ij}^t(k_+^t) + \mathbf{w}^t(k_-^t)^\top \mathbf{x}_{ij}^t(k_-^t) + b_{ij}^t \quad (16)$$

where $\triangle \mathbf{x}_{ij}^t(k_+^t)$ is the ring boundary of $\mathbf{x}_{ij}^t(k_+^t)$ in the input image/feature map with zeros inside and $\mathbf{x}_{ij}^t(k_-^t)$ is padded with zeros.

Hence, the partial derivative of the activation $y_{ij}^t$ w.r.t. the upper-bound filter $\mathbf{w}^t(k_+^t)$ can be calculated as follows:
$$\frac{\partial y_{ij}^t}{\partial \mathbf{w}^t(k_+^t)} = \mathbf{x}_{ij}^t(k_-^t)^\top + \alpha^t \triangle \mathbf{x}_{ij}^t(k_+^t)^\top \quad (17)$$

With the chain rule, the derivative of CNN loss w.r.t. $\mathbf{w}^t(k_+)$ can be calculated as
$$\frac{\partial L^t}{\partial \mathbf{w}^t(k_+^t)} = \sum_{i,j} \frac{\partial L^t}{\partial y_{ij}^t} \frac{\partial y_{ij}^t}{\partial \mathbf{w}^t(k_+^t)} \quad (18)$$

Thus, the upper-bound filter $\mathbf{w}(k_+)$ can be updated iteratively using the SGD strategy. As a result, the filter $\mathbf{w}(k)$ with a continuous size $k$ can be updated as in Eq. 5.

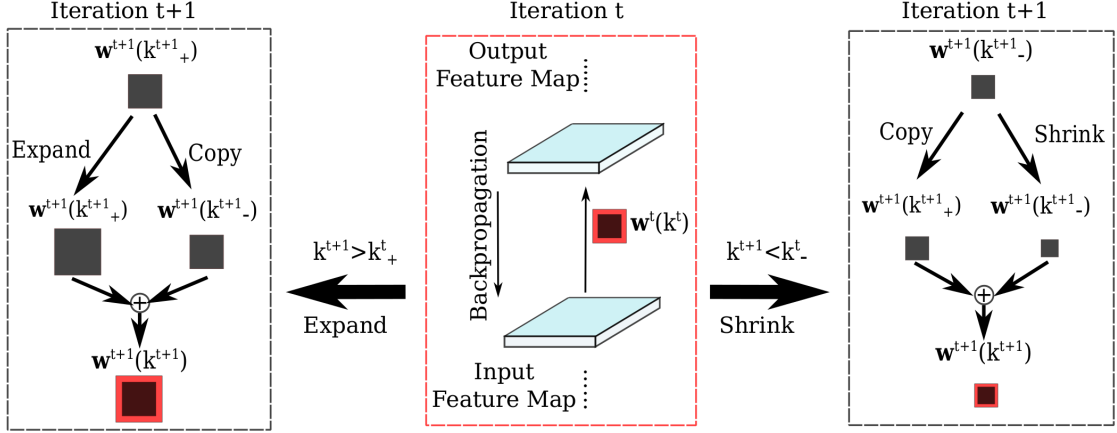

Figure 3. When the filter size $k$ is updated during backpropagation, it may be out of the interval $[k_-^t, k_+^t)$. In this case, transformation operations are needed to update the sizes of the upper-bound and lower-bound filters after updating their coefficients. Specifically, an expanding operation is employed to increase the sizes of both upper-bound and lower-bound filters; whereas a shrinking operation is used to decrease the filter sizes.

**Transforming the upper-bound and lower-bound filters:**
According to Eq. 15, the filter size $k$ can be continuously updated over time. As long as $k^{t+1}$ is in the interval of $[k_-^t, k_+^t)$, the upper-bound and lower bound filters remain the same sizes as those in the $t^{th}$ iteration, i.e., $k_-^{t+1} = k_-^t$ and $k_+^{t+1} = k_+^t$. However, as the filter size $k$ is updated, $k^{t+1}$ may be outside of the interval of $[k_-^t, k_+^t)$. Consequently, both the sizes of the upper-bound and lower-bound filters should be updated. As illustrated in Fig. 3, we define *transformation operations*, including $expanding$ and $shrinking$ to update the upper-bound and lower-bound filters to accommodate a size change.

Note that, the transformation operations are conducted after updating coefficients of the upper-bound and lower-bound filters.

*Expanding:* When $k^{t+1} > k_+^t$, the upper-bound and lower-bound filters $\mathbf{w}^{t+1}(k_+^{t+1})$ and $\mathbf{w}^{t+1}(k_-^{t+1})$ should be updated by an expanding operation as follows:

$$\mathbf{w}^{t+1}(k_-^{t+1}) = \mathbf{w}^{t+1}(k_+^{t+1})$$
$$\mathbf{w}^{t+1}(k_+^{t+1}) = expand(\mathbf{w}^{t+1}(k_+^{t+1})) \quad (19)$$

where $expand(\cdot)$ is a function to increase the filter size, particularly by padding values from the nearest neighbors of the original filter as illustrated in Figure 4.

*Shrinking:* As opposed to the $expand(\cdot)$ function, when $k^{t+1} < k_-^t$, the upper-bound and lower-bound filters $\mathbf{w}^{t+1}(k_-^{t+1})$ and $\mathbf{w}^{t+1}(k_+^{t+1})$ will be shrunk as follows:

$$\mathbf{w}^{t+1}(k_+^{t+1}) = \mathbf{w}^{t+1}(k_-^{t+1})$$
$$\mathbf{w}^{t+1}(k_-^{t+1}) = shrink(\mathbf{w}^{t+1}(k_-^{t+1})) \quad (20)$$

where $shrink(\cdot)$ is a function to decrease the filter size, specifically by filling the boundary with zeros as shown in Figure 4.

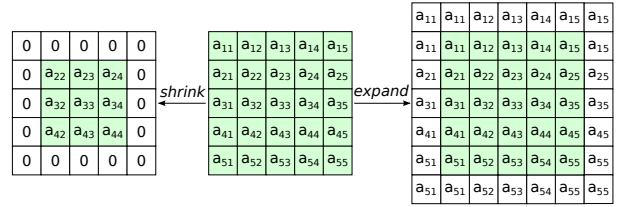

Figure 4. An illustration of the *shrink* and *expand* operations to change the filter size. The *shrink* operation sets zeros to the outside boundary; while the *expand* operation is to pad the outside boundary with the nearest neighbors from the original filter.

**Remark 5.** *There are alternative methods that can be used to expand or shrink the filters. For example, we have also tried to resize the filter by bicubic interpolation. However, the recognition performance became worse. The reason is that the filters learned in the previous iterations are distorted after scaling and thus, may fail to activate the patterns in the images. In contrast, the proposed expand and shrink functions can well preserve the learned filters.*

**Updating other parameters:** In addition to updating the filter size $k$ and the convolution filter $\mathbf{w}(k)$, we should also update the bias $b_{ij}$ and the feature $\mathbf{x}_{ij}$ during backpropagation. Based on the forward activation function as defined in Eq. 6, the derivative of feature activation $y_{ij}^t$ w.r.t. $\mathbf{x}_{ij}^t(k_+^t)$ can be calculated as below:

$$\frac{\partial y_{ij}^t}{\partial \mathbf{x}_{ij}^t(k_+^t)} = \mathbf{w}^t(k^t) \quad (21)$$

With the chain rule, the derivative of CNN loss w.r.t. $\mathbf{x}_{ij}^t(k_+^t)$ can be calculated as:

$$\frac{\partial L^t}{\partial \mathbf{x}_{ij}^t(k_+^t)} = \frac{\partial L^t}{\partial y_{ij}^t} \frac{\partial y_{ij}^t}{\partial \mathbf{x}_{ij}^t(k_+^t)} \quad (22)$$

**Algorithm 1** The forward-backward propagation algorithm for the OFS-CNN

**Input:** Input images or feature maps from the previous layer **x** and an initial filter size $k^0 \in \mathbb{R}^+$.
**Initialization:**
Initialize $k_+^0$ and $k_-^0$ as Eq. 2.
Randomly initialize the convolution filter $\mathbf{w}^0(k_+^0)$.
**for** iteration $t$ from 0 to $T$ **do**
  //Forward:
  $\mathbf{w}^t(k_-^t) = shrink(\mathbf{w}^t(k_+^t))$
  Calculate the convolution filter $\mathbf{w}^t(k^t)$ based on Eq. 5
  Calculate the forward activation $y_{ij}(k)$ based on Eq. 6
  //Backward:
  Calculate the derivative of activation w.r.t. $k^t$, $\mathbf{w}^t(k_+^t)$, and $\mathbf{x}^t$, based on Eqs.13, 17, and 21, respectively
  Calculate the derivative of loss w.r.t. $k^t$, $\mathbf{w}^t(k_+^t)$, and $\mathbf{x}^t$, based on Eqs.14, 18, and 22, respectively
  Update $k^{t+1}$, $\mathbf{w}^{t+1}(k_+^{t+1})$, and $\mathbf{x}^{t+1}$ based on SGD
  Update the bias using standard CNN backpropagation
  //Transformation:
  **if** $k^{t+1} > k_+^t$ **then**
    $k_-^{t+1} = k_+^t$
    $k_+^{t+1} = k_+^t + 2$
    Expand the upper-bound and lower bound filters $\mathbf{w}^{t+1}(k_+^{t+1})$ and $\mathbf{w}^{t+1}(k_-^{t+1})$ as in Eq. 19
  **else if** $k^{t+1} < k_-^t$ **then**
    $k_+^{t+1} = k_-^t$
    $k_-^{t+1} = k_-^t - 2$
    Shrink the upper-bound and lower bound filters $\mathbf{w}^{t+1}(k_+^{t+1})$ and $\mathbf{w}^{t+1}(k_-^{t+1})$ as in Eq. 20
  **end if**
**end for**

Hence, the feature $\mathbf{x}_{ij}$ can be updated using the SGD strategy and will be further backpropagated to update the parameters in the lower layers. The backpropagation of $b_{ij}^t$ is exactly the same as that in the traditional CNNs. The forward and backward propagation process for the proposed OFS-CNN is summarized in Algorithm 1.

## 4. Experiments

To demonstrate the effectiveness of the proposed model, extensive experiments have been conducted on two benchmark AU-coded databases, i.e., the BP4D database [26] and the DISFA database [20], containing spontaneous facial behavior with moderate head movements. Specifically, the BP4D database [26] has 11 AUs and 41 subjects with 146,847 images; and the DISFA database [20] has 12 AUs and 27 subjects with 130,814 images. Following the experimental setup of the state-of-the-art methods (DRML [34] and PL-CNN [28]), two AUs, i.e., AU5 and AU20, which appear less than 5% of the frames in the DISFA database, are not considered in the experiments.

### 4.1. Pre-Processing

First, facial landmarks are detected, from which face alignment can be conducted to reduce the variations from scaling and in-plane rotation. For the DISFA database [20], 66 landmarks are detected using a state-of-the-art method [1]. For the BP4D database [26], the 49 landmarks provided in the database are used for face alignment. Based on the extracted facial landmarks, face regions are aligned based on three fiducial points: the centers of the two eyes and the mouth, and then scaled to $64 \times 48$ [1]. Following the work [10], each face image is warped to a frontal view to reduce variations from face pose; and then sequence normalization is performed by subtracting the mean and dividing the standard deviation calculated from the video sequence to reduce the identity-related information and to enhance appearance and geometrical changes caused by AUs.

### 4.2. CNN Implementation Details

The proposed OFS-CNN is modified from cifar10_quick in Caffe [13], which consists of three convolutional layers, two average pooling layers, two FC layers, and ending with the weighted sigmoid cross entropy loss layer for calculating the loss. Specifically, all the convolutional layers have a stride of 1. The first two convolutional layers have 32 filters, whose output feature maps are sent to a rectified layer followed by the average pooling layer with a downsampling stride of 3. The last convolutional layer has 64 filters, whose output feature maps are fed into an FC layer with 64 nodes. Finally, the output of the last FC layer, which contains a single node, is sent to the loss layer. The SGD, with a momentum of 0.9 and a mini-batch size of 100, is used for training the CNN. Each AU has one trained CNN model.

All filter sizes are $5 \times 5$ in the original cifar10_quick [13] and will be used for the baseline CNN for comparison. In the OFS-CNN, all filter sizes are initialized with 4, implying $\alpha^0 = 0.5$, $k_+^0 = 5$, and $k_-^0 = 3$.

### 4.3. Experimental Results

The proposed OFS-CNN is compared with the baseline CNN with fixed convolution filter sizes on the two benchmark datasets. Since the BP4D database [26] provides the training and development partitions, an average performance of five runs is reported to reduce the influence of the randomness during training. For the DISFA database [20], a 9-fold cross-validation strategy is employed, such that the training and testing subjects are mutually exclusive. Experimental results are reported in terms of the average F1 score and 2AFC score (area under ROC curve).

---
[1] In the experiments, three resolutions, i.e., $64 \times 48$, $128 \times 96$, $256 \times 192$ are employed to evaluate the proposed OFS-CNN on different resolutions.

Table 1. Performance comparison of the proposed OFS-CNNs and traditional CNNs with varying filter size on the BP4D database [26]. In the *1-layer OFS-CNN*, the filter size is learned only for the first layer. The average converged filter size is reported for each AU, respectively. The results are calculated from 5 runs in terms of the average F1 score and the 2AFC score. The underline highlights the best performance among the 4 fixed filter sizes. The **bold** highlights the best performance among all models.

| AUs | CNN-Filter3 | | CNN-Filter5 | | CNN-Filter7 | | CNN-Filter9 | | 1-layer OFS-CNN | | | 3-layer OFS-CNN | | |
|---|---|---|---|---|---|---|---|---|---|---|---|---|---|---|
| | F1 | 2AFC | F1 | 2AFC | F1 | 2AFC | F1 | 2AFC | F1 | 2AFC | Converged Size | F1 | 2AFC | Converged Size |
| AU1 | 0.315 | 0.577 | 0.313 | 0.578 | 0.310 | 0.577 | 0.315 | 0.583 | 0.320 | 0.586 | 6.0 | **0.348** | **0.628** | 5.2, 5.1, 5.1 |
| AU2 | 0.291 | 0.591 | 0.277 | 0.573 | 0.284 | 0.586 | 0.279 | 0.575 | 0.291 | 0.592 | 5.8 | **0.312** | **0.626** | 5.2, 5.3, 4.9 |
| AU4 | 0.362 | 0.654 | 0.358 | 0.649 | 0.361 | 0.653 | 0.367 | 0.661 | 0.362 | 0.661 | 6.0 | **0.376** | **0.673** | 5.1, 5.5, 4.8 |
| AU6 | 0.677 | 0.754 | 0.693 | 0.775 | 0.688 | 0.771 | 0.689 | 0.773 | 0.685 | 0.764 | 6.0 | **0.723** | **0.811** | 5.1, 4.7, 4.7 |
| AU7 | 0.640 | 0.654 | 0.643 | 0.658 | 0.652 | **0.661** | 0.646 | 0.659 | **0.658** | 0.660 | 6.0 | 0.634 | 0.652 | 5.0, 4.8, 4.7 |
| AU10 | 0.706 | 0.720 | 0.726 | 0.728 | 0.716 | 0.720 | 0.711 | 0.723 | 0.720 | 0.725 | 6.0 | **0.739** | **0.758** | 4.6, 5.1, 4.8 |
| AU12 | 0.749 | 0.786 | 0.763 | 0.805 | 0.759 | 0.805 | 0.750 | 0.791 | 0.768 | 0.801 | 6.1 | **0.799** | **0.855** | 4.8, 5.9, 4.8 |
| AU14 | 0.505 | 0.582 | 0.517 | 0.597 | 0.525 | 0.600 | 0.523 | 0.593 | 0.521 | 0.600 | 5.4 | **0.532** | **0.635** | 5.1, 4.6, 4.5 |
| AU15 | 0.298 | 0.603 | 0.296 | 0.599 | 0.306 | **0.611** | 0.316 | 0.622 | 0.305 | 0.609 | 6.0 | 0.300 | 0.607 | 5.2, 4.9, 4.8 |
| AU17 | 0.547 | 0.676 | 0.550 | 0.683 | 0.553 | 0.683 | 0.544 | 0.678 | 0.532 | 0.673 | 5.7 | 0.542 | **0.694** | 4.9, 4.7, 4.6 |
| AU23 | 0.337 | 0.651 | 0.348 | 0.658 | 0.352 | 0.657 | 0.350 | **0.659** | 0.345 | 0.655 | 6.1 | **0.355** | **0.659** | 5.4, 4.6, 4.7 |
| AVE | 0.493 | 0.659 | 0.499 | 0.664 | 0.501 | 0.666 | 0.499 | 0.665 | 0.501 | 0.666 | 5.9 | **0.515** | **0.691** | 5.0, 5.0, 4.8 |

**Exhaustive search vs optimization of filter size:** We first show that the proposed OFS-CNN is capable of learning the optimal filter sizes. Specifically, baseline CNNs are designed with varying filter sizes including $3 \times 3$, $5 \times 5$, $7 \times 7$, and $9 \times 9$ in the first convolutional layer. In addition to the *3-layer OFS-CNN*, where the filter sizes in all three convolutional layers are learned, a *1-layer OFS-CNN* is designed where the filter size is learned only for the first layer. All the baseline CNNs and the *1-layer OFS-CNN* used the fixed filter sizes ($5 \times 5$) for the other two convolutional layers. All the models in comparison are trained on the training partition and tested on the development partition of the BP4D database [26]. The results are reported in Table 1, which are calculated as the average of 5 runs. The average filter size of OFS-CNNs is reported for each AU at the $2000^{th}$ iteration, where most of the CNN models are converged in our experiments.

As shown in Table 1, the *1-layer OFS-CNN* not only outperforms *CNN-Filter5* as in the original cifar10_quick [13] in terms of the average F1 score (0.501 vs 0.499) and the average 2AFC score (0.666 vs 0.664), but also achieves similar performance as *CNN-Filter7* that has the best performance among all baseline CNNs. Furthermore, the *3-layer OFS-CNN* beats all models compared to in terms of the average F1 score and 2AFC score. This demonstrates that the proposed OFS-CNN is superior to the best CNN model obtained by exhaustive search. In addition, the learned filter size is often consistent with the best filter size obtained by exhaustive search, which is either the upper-bound or lower-bound filter size in the OFS-CNN.

**OFS-CNNs on different image resolutions:** We also show that the learned filter sizes adapt well to changes in image resolutions. Specifically, experiments have been conducted to compare the proposed OFS-CNN and the baseline CNN on the BP4D database [26] with different resolutions of the input images. All the CNN models have similar CNN structure as described in Section 4.2. In order to accommodate the changes in the resolution, the number of nodes in the first FC layer is set to 64, 128, and 256 for resolutions of $64 \times 48$, $128 \times 96$, and $256 \times 192$, respectively, for all models in comparison. In this set of experiments, the *3-layer OFS-CNN* is employed and the average converged filter sizes for each AU under each resolution are reported in Table 3.

Table 2. Performance comparison of the proposed OFS-CNN and the baseline CNN for varying image resolutions on the BP4D database [26] in terms of the average F1 score. The **bold** highlights the best performance among all models.

| Resolution | $64 \times 48$ | | $128 \times 96$ | | $256 \times 192$ | |
|---|---|---|---|---|---|---|
| Layer | CNN | OFS-CNN | CNN | OFS-CNN | CNN | OFS-CNN |
| AU1 | 0.313 | 0.348 | 0.340 | 0.345 | 0.332 | **0.416** |
| AU2 | 0.277 | **0.312** | 0.307 | 0.303 | 0.278 | 0.305 |
| AU4 | 0.358 | 0.376 | 0.411 | **0.415** | 0.324 | 0.391 |
| AU6 | 0.693 | 0.723 | 0.721 | 0.729 | 0.676 | **0.745** |
| AU7 | 0.643 | 0.634 | 0.642 | **0.649** | 0.504 | 0.628 |
| AU10 | 0.726 | 0.739 | 0.718 | **0.754** | 0.690 | 0.743 |
| AU12 | 0.763 | 0.799 | 0.774 | 0.805 | 0.697 | **0.812** |
| AU14 | 0.517 | 0.532 | 0.552 | **0.562** | 0.544 | 0.555 |
| AU15 | 0.296 | 0.300 | 0.331 | **0.337** | 0.323 | 0.326 |
| AU17 | 0.550 | 0.542 | 0.561 | 0.563 | 0.540 | **0.568** |
| AU23 | 0.348 | 0.355 | 0.381 | 0.398 | 0.354 | **0.413** |
| AVE | 0.499 | 0.515 | 0.522 | 0.533 | 0.478 | **0.537** |

Table 3. The average converged filter sizes for varying image resolutions on the BP4D database [26]. The **bold** highlights the filter sizes with the best performance.

| Resolution | $64 \times 48$ | | | $128 \times 96$ | | | $256 \times 192$ | | |
|---|---|---|---|---|---|---|---|---|---|
| Layer | conv1 | conv2 | conv3 | conv1 | conv2 | conv3 | conv1 | conv2 | conv3 |
| AU1 | 5.2 | 5.1 | 5.1 | 5.5 | 4.9 | 5.1 | **6.2** | **4.9** | **4.9** |
| AU2 | **5.2** | **5.3** | **4.9** | 6.0 | 4.8 | 4.9 | 5.9 | 5.3 | 5.1 |
| AU4 | 5.1 | 5.5 | 4.8 | **5.7** | **5.8** | **5.7** | 5.8 | 5.8 | 5.8 |
| AU6 | 5.1 | 4.7 | 4.7 | 5.4 | 4.7 | 4.7 | **5.7** | **4.8** | **4.8** |
| AU7 | 5.0 | 4.8 | 4.7 | **5.3** | **4.6** | **4.7** | 5.6 | 4.8 | 4.8 |
| AU10 | 4.6 | 5.1 | 4.8 | **5.5** | **4.8** | **4.8** | 5.5 | 5.5 | 4.9 |
| AU12 | 4.8 | 5.9 | 4.8 | 5.2 | 5.5 | 5.9 | **5.7** | **5.5** | **5.4** |
| AU14 | 5.1 | 4.6 | 4.5 | **5.3** | **4.6** | **4.6** | 5.9 | 4.6 | 4.5 |
| AU15 | 5.2 | 4.9 | 4.8 | **5.5** | **4.8** | **4.8** | 5.5 | 4.8 | 4.8 |
| AU17 | 4.9 | 4.7 | 4.6 | 5.6 | 4.5 | 4.5 | **5.3** | **4.6** | **4.5** |
| AU23 | 5.4 | 4.6 | 4.7 | 6.0 | 4.7 | 4.7 | **5.9** | **4.8** | **4.7** |
| AVE | 5.0 | 5.0 | 4.8 | 5.5 | 5.0 | 5.0 | 5.7 | 5.0 | 5.0 |

As shown in Table 2, most of AUs prefer a higher image resolution to preserve subtle cues of facial appearance changes. However, the performance of the baseline CNN decreases for the highest resolution $256 \times 192$. When the image resolution increases, the receptive field covers a smaller actual area of the whole face when using the same $5 \times 5$ filter size, compared to lower resolutions. In contrast, the proposed OFS-CNN can optimize filter size at various image resolutions. As shown in Table 3, the OFS-CNN has the largest average filter size of 5.7 for conv1 (the first convolutional layer) for $256 \times 192$ and thus, can benefit from an increased receptive field because of the $7 \times 7$ upper-bound filter. As a result, the OFS-CNN outperforms the baseline CNN for all image resolutions, especially for $256 \times 192$ by 6%, in terms of the average F1 score.

**Comparison with the CNNs using multiple filter sizes:** We also compare the proposed *3-layer OFS-CNN* to the CNN structure with multiple filter sizes, i.e., the inception module [23]. In particular, the GoogLeNet [23] with 7 inception modules is trained and evaluated on the BP4D database with an image resolution of $240 \times 240$.

Table 4. Comparison with the GoogLeNet on the BP4D database in terms of F1 score.

| AUs | % | GoogLeNet | OFS-CNN $128 \times 96$ | OFS-CNN $256 \times 192$ |
|---|---|---|---|---|
| AU1 | 23.1 | 0.369 | 0.345 | **0.416** |
| AU2 | 17.9 | 0.267 | 0.303 | **0.305** |
| AU4 | 22.7 | **0.498** | 0.415 | 0.391 |
| AU6 | 46.0 | **0.746** | 0.729 | 0.745 |
| AU7 | 52.6 | **0.657** | 0.649 | 0.628 |
| AU10 | 59.6 | **0.768** | 0.754 | 0.743 |
| AU12 | 55.8 | **0.836** | 0.805 | 0.812 |
| AU14 | 52.1 | 0.503 | **0.562** | 0.555 |
| AU15 | 18.0 | 0.325 | **0.337** | 0.326 |
| AU17 | 32.6 | 0.511 | 0.563 | **0.568** |
| AU23 | 17.0 | 0.376 | 0.398 | **0.413** |
| AVE | - | 0.531 | 0.533 | **0.537** |

As shown in Table 4, the OFS-CNN with a shallow structure (15 layers, trained in 3,000 iterations) performs noticeably better than the GoogLeNet (100 layers, trained in 20,000 iterations) in terms of the average F1 score. The improvement becomes more substantial for the AUs with a lower occurrence rate such as AU2 (17.9%) and AU23 (17.0%). Furthermore, the GoogLeNet is much more complex compared to the OFS-CNN and thus, demands more training data. Note that the proposed OFS-CNN runs more than 8 times faster on a $128 \times 96$ image and more than 6 times faster on a $256 \times 192$ image than the GoogLeNet ($240 \times 240$) during testing, which is critical and hence, highly desirable for real-time applications.

**Comparison with the baseline CNN on the DISFA database [20]:** As illustrated in Table 5, the proposed OFS-CNN also outperforms the baseline CNN with a notable margin in terms of the average F1 score on the DISFA database [20]. The experiments are conducted on the image resolution of $128 \times 96$ using the *3-layer OFS-CNN*.

**Comparison with state-of-the-art methods:** In addition to the baseline CNN, we further compare the proposed OFS-CNN with state-of-the-art methods, particularly the most re-cent approaches based on CNNs [8, 9, 34, 28], on the two benchmark databases. As shown in Table 6, the proposed OFS-CNN achieves the state-of-the-art performance of AU recognition on the two databases [2].

Table 5. Performance comparison with the baseline CNN on the DISFA database [20] in terms of the average F1 score and the 2AFC score.

| AUs | CNN (baseline) | | OFS-CNN | |
|---|---|---|---|---|
| | F1 | 2AFC | F1 | 2AFC |
| AU1 | 0.321 | 0.778 | **0.437** | **0.833** |
| AU2 | **0.424** | **0.865** | 0.400 | 0.812 |
| AU4 | 0.567 | 0.833 | **0.672** | **0.862** |
| AU6 | **0.610** | **0.896** | 0.590 | **0.896** |
| AU9 | 0.417 | **0.876** | **0.497** | 0.873 |
| AU12 | **0.786** | 0.950 | 0.758 | **0.956** |
| AU15 | 0.298 | 0.794 | **0.378** | **0.799** |
| AU17 | 0.452 | **0.831** | **0.523** | 0.823 |
| AU25 | 0.716 | 0.847 | **0.724** | **0.849** |
| AU26 | **0.564** | **0.827** | 0.548 | 0.800 |
| AVE | 0.515 | **0.850** | **0.553** | **0.850** |

Table 6. Performance comparison with the state-of-the-art CNN based methods on the BP4D and the DISFA databases in terms of F1 score and 2AFC score.

| BP4D | | | DISFA | | |
|---|---|---|---|---|---|
| Methods | F1 | 2AFC | Methods | F1 | 2AFC |
| DL [9] | 0.522 | N/A | ML-CNN [8] | N/A | 0.757 |
| AlexNet [34] | 0.384 | 0.422 | AlexNet [34] | 0.236 | 0.491 |
| LCN [34] | 0.466 | 0.544 | LCN [34] | 0.240 | 0.468 |
| ConvNet [34] | 0.470 | 0.518 | ConvNet [34] | 0.231 | 0.458 |
| DRML [34] | 0.483 | 0.560 | DRML [34] | 0.267 | 0.523 |
| PL-CNN [28] | 0.491 | N/A | PL-CNN [28] | 0.584 | N/A |
| **OFS-CNN** | 0.537 | 0.722 | **OFS-CNN** | 0.553 | 0.850 |

## 5. Conclusion and Future Work

Traditional CNNs have a predefined and fixed integral filter size for each convolutional layer, which may be not optimal for all tasks as well as for all image resolutions. In this work, we proposed a novel OFS-CNN with a forward-backward propagation algorithm to iteratively optimize the filter size while learning the convolution filters. Upper-bound and lower-bound filters are defined to facilitate the convolution operations with continuous-size filters; and transformation operations are developed to accommodate the size changes of the filters. Experimental results on two benchmark AU-coded spontaneous databases have shown that the OFS-CNN outperforms the baseline CNNs with the best filter size found by exhaustive search and achieves better or at least comparable performance to the state-of-the-art CNN-based methods. Furthermore, the OFS-CNN has been shown to be effective for automatically adapting filter sizes to different image resolutions. In the current practice, different channels of a single convolutional layer share a single filter size. In the future, the OFS-CNN will be extended to learn a filter size for each channel, which would be more effective for learning variously sized patterns.

---

[2]The performance of the ML-CNN was reported for 10 AUs on the DISFA database [20].

# References

[1] A. Asthana, S. Zafeiriou, S. Cheng, and M. Pantic. Robust discriminative response map fitting with constrained local models. In *CVPR*, pages 3444–3451, 2013.

[2] T. Baltrusaitis, M. Mahmoud, and P. Robinson. Cross-dataset learning and person-specific normalisation for automatic action unit detection. In *FG*, volume 6, pages 1–6, 2015.

[3] M. S. Bartlett, G. Littlewort, M. G. Frank, C. Lainscsek, I. Fasel, and J. R. Movellan. Recognizing facial expression: Machine learning and application to spontaneous behavior. In *CVPR*, pages 568–573, 2005.

[4] K. Chatfield, K. Simonyan, A. Vedaldi, and A. Zisserman. Return of the devil in the details: Delving deep into convolutional nets. In *BMVC*, 2014.

[5] A. Dhall, O. Ramana Murthy, R. Goecke, J. Joshi, and T. Gedeon. Video and image based emotion recognition challenges in the wild: Emotiw 2015. In *ICMI*, pages 423–426. ACM, 2015.

[6] P. Ekman, W. V. Friesen, and J. C. Hager. *Facial Action Coding System: the Manual*. Research Nexus, Div., Network Information Research Corp., Salt Lake City, UT, 2002.

[7] B. Fasel. Head-pose invariant facial expression recognition using convolutional neural networks. In *ICMI*, pages 529–534, 2002.

[8] S. Ghosh, E. Laksana, S. Scherer, and L. Morency. A multi-label convolutional neural network approach to cross-domain action unit detection. *ACII*, 2015.

[9] A. Gudi, H. E. Tasli, T. M. den Uyl, and A. Maroulis. Deep learning based FACS action unit occurrence and intensity estimation. In *FG*, 2015.

[10] S. Han, Z. Meng, S. KHAN, and Y. Tong. Incremental boosting convolutional neural network for facial action unit recognition. In D. D. Lee, M. Sugiyama, U. V. Luxburg, I. Guyon, and R. Garnett, editors, *NIPS*, pages 109–117. Curran Associates, Inc., 2016.

[11] K. He, X. Zhang, S. Ren, and J. Sun. Deep residual learning for image recognition. In *CVPR*, pages 770–778, 2016.

[12] S. Jaiswal and M. F. Valstar. Deep learning the dynamic appearance and shape of facial action units. In *WACV*, 2016.

[13] Y. Jia, E. Shelhamer, J. Donahue, S. Karayev, J. Long, R. Girshick, S. Guadarrama, and T. Darrell. Caffe: Convolutional architecture for fast feature embedding. In *ACM MM*, pages 675–678. ACM, 2014.

[14] B. Jiang, B. Martinez, M. F. Valstar, and M. Pantic. Decision level fusion of domain specific regions for facial action recognition. In *ICPR*, pages 1776–1781, 2014.

[15] H. Jung, S. Lee, J. Yim, S. Park, and J. Kim. Joint fine-tuning in deep neural networks for facial expression recognition. In *ICCV*, pages 2983–2991, 2015.

[16] T. W. Ke, M. Maire, and X. Y. Stella. Multigrid neural architectures. 2017.

[17] B.-K. Kim, H. Lee, J. Roh, and S.-Y. Lee. Hierarchical committee of deep cnns with exponentially-weighted decision fusion for static facial expression recognition. In *ICMI*, pages 427–434, 2015.

[18] G. Levi and T. Hassner. Age and gender classification using convolutional neural networks. In *CVPR Workshops*, pages 34–42, 2015.

[19] M. Liu, S. Li, S. Shan, R. Wang, and X. Chen. Deeply learning deformable facial action parts model for dynamic expression analysis. In *ACCV*, 2014.

[20] S. M. Mavadati, M. H. Mahoor, K. Bartlett, P. Trinh, and J. F. Cohn. Disfa: A spontaneous facial action intensity database. *IEEE Trans. on Affective Computing*, 4(2):151–160, 2013.

[21] H.-W. Ng, V. D. Nguyen, V. Vonikakis, and S. Winkler. Deep learning for emotion recognition on small datasets using transfer learning. In *ICMI*, pages 443–449, 2015.

[22] S. Saxena and J. Verbeek. Convolutional neural fabrics. In *Advances in Neural Information Processing Systems*, pages 4053–4061, 2016.

[23] C. Szegedy, W. Liu, Y. Jia, P. Sermanet, S. Reed, D. Anguelov, D. Erhan, V. Vanhoucke, and A. Rabinovich. Going deeper with convolutions. In *CVPR*, pages 1–9, 2015.

[24] Y. Tang. Deep learning using linear support vector machines. In *ICML*, 2013.

[25] Z. Tősér, L. Jeni, A. Lőrincz, and J. Cohn. Deep learning for facial action unit detection under large head poses. In *ECCV*, pages 359–371. Springer, 2016.

[26] M. Valstar, J. Girard, T. Almaev, G. McKeown, M. Mehu, L. Yin, M. Pantic, and J. Cohn. FERA 2015 - second facial expression recognition and analysis challenge. *FG*, 2015.

[27] M. F. Valstar, M. Mehu, B. Jiang, M. Pantic, and K. Scherer. Meta-analysis of the first facial expression recognition challenge. *IEEE T-SMC-B*, 42(4):966–979, 2012.

[28] S. Wu, S. Wang, B. Pan, and Q. Ji. Deep facial action unit recognition from partially labeled data. In *ICCV*, pages 3951–3959, 2017.

[29] P. Yang, Q. Liu, and M. D. N. Boosting encoded dynamic features for facial expression recognition. *Pattern Recognition Letters*, 30(2):132–139, Jan. 2009.

[30] Z. Yu and C. Zhang. Image based static facial expression recognition with multiple deep network learning. In *ICMI*, pages 435–442, 2015.

[31] A. Yuce, H. Gao, and J. Thiran. Discriminant multi-label manifold embedding for facial action unit detection. In *FG*, 2015.

[32] M. D. Zeiler and R. Fergus. Visualizing and understanding convolutional networks. In *ECCV*, pages 818–833, 2014.

[33] G. Zhao and M. Pietiäinen. Dynamic texture recognition using local binary patterns with an application to facial expressions. *IEEE T-PAMI*, 29(6):915–928, June 2007.

[34] K. Zhao, W. Chu, and H. Zhang. Deep region and multi-label learning for facial action unit detection. In *CVPR*, pages 3391–3399, 2016.